\documentclass{bmvc2k}

\title{Unsupervised Domain Adaptive Fundus Image Segmentation with Few Labeled Source Data}

\addauthor{Qianbi Yu}{qiyu2312@uni.sydney.edu.au}{1}
\addauthor{Dongnan Liu}{dongnan.liu@sydney.edu.au}{1}
\addauthor{Chaoyi Zhang}{czha5168@uni.sydney.edu.au}{1}
\addauthor{Xinwen Zhang}{xzha9145@uni.sydney.edu.au}{1}
\addauthor{Weidong Cai}{tom.cai@sydney.edu.au}{1}

\addinstitution{
    School of Computer Science\\
    University of Sydney\\
    Sydney, Australia
}
\runninghead{YU ET AL.}{UNSUPERVISED DOMAIN ADAPTATION WITH FEW SHOT DATA}

\begin{document}

\maketitle

\begin{abstract}
Deep learning-based segmentation methods have been widely employed for automatic glaucoma diagnosis and prognosis. In practice, fundus images obtained by different fundus cameras vary significantly in terms of illumination and intensity. Although recent unsupervised domain adaptation (UDA) methods enhance the models' generalization ability on the unlabeled target fundus datasets, they always require sufficient labeled data from the source domain, bringing auxiliary data acquisition and annotation costs. To further facilitate the data efficiency of the cross-domain segmentation methods on the fundus images, we explore UDA optic disc and cup segmentation problems using few labeled source data in this work. We first design a Searching-based Multi-style Invariant Mechanism to diversify the source data style as well as increase the data amount. Next, a prototype consistency mechanism on the foreground objects is proposed to facilitate the feature alignment for each kind of tissue under different image styles. Moreover, a cross-style self-supervised learning stage is further designed to improve the segmentation performance on the target images. Our method has outperformed several state-of-the-art UDA segmentation methods under the UDA fundus segmentation with few labeled source data. 
\end{abstract}

%-------------------------------------------------------------------------
\section{Introduction}
\label{sec:intro}
Glaucoma is a chronic eye condition that causes progressive damage to the optic nerve and eventually leads to blindness if left untreated~\cite{almazroa2015optic}. In clinical practice, accurate examination of the head of the optic nerve i.e. cup-to-disc ratio is crucial for early detection and treatment of glaucoma diagnosis~\cite{meng2021bi}. Recently, deep learning-based models have been widely used for automatic optic cup and disc segmentation in fundus images~\cite{sevastopolsky2017optic,fu2018joint} and achieved appealing performance. Nevertheless, these methods will suffer from performance drop when validated on new datasets with unseen distributions due to the domain shift issue~\cite{ganin2015unsupervised,patel2015visual}. To this end, unsupervised domain adaptation (UDA) methods have been proposed to enhance the models' generalization ability, by transferring the knowledge from the labeled source data to the unlabeled target data~\cite{madani2018semi,javanmardi2018domain,gholami2018novel,liu2020unsupervised,liu2020pdam,li2022domain}. 

Recently, several methods have been proposed to further facilitate the data-efficiency of the UDA medical image segmentation~\cite{Li2021FewShotDA,chen2021source,zhao2021mt}.~\cite{Li2021FewShotDA} proposed to train the UDA model with few target images, and~\cite{chen2021source} explored a source-free UDA setting without sharing the raw source data with the target domain. However, they still require the model to be optimized with sufficient labeled source images, which might be challenging to fulfill in practical applications. Acquiring the pixel-level annotations for the optic discs and cups is time-consuming and error-prone for automatic fundus segmentation datasets~\cite{wang2019boundary}. On the other side, directly training the UDA models with insufficient labeled source data can easily cause over-fitting problems and limit the segmentation performance. To further reduce the amount of data required and maintain high-performance cross-domain segmentation, we explore the UDA fundus image segmentation problem with few labeled source data. Although~\cite{zhao2021mt} also explores a similar problem as ours via a teacher-student framework, they ignore the category relationship of the foreground tissues. Given the extremely insufficient supervision (e.g., given no more than 10\% of the labeled training data), the model would suffer from misalignment due to the lack of semantic-level information learning.

%-------------------------------------------------------------------------

\begin{figure}
\centering
\includegraphics[width=0.7\textwidth]{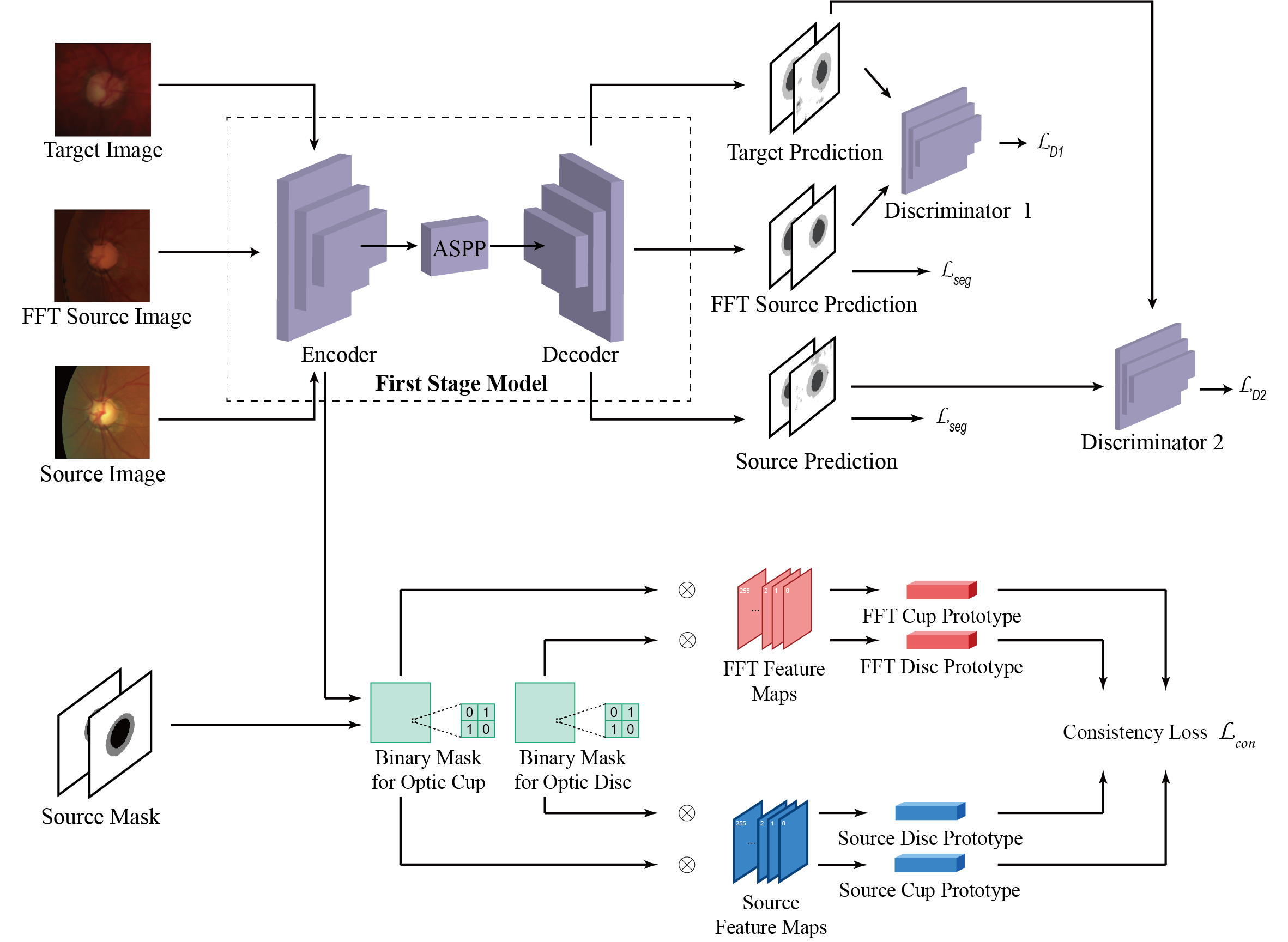}\\
\caption{Overview of the proposed framework for cross-domain optic disc and cup segmentation with few labeled data. The first stage model is trained using target and synthesis images with additional discriminative and consistency loss. The customized consistency loss utilizes the mask pooling technique and computes the cosine similarity of the class-prototype vectors.}
\label{fig:overview}
\end{figure}

%-------------------------------------------------------------------------
In this work, a novel framework is proposed to transfer the knowledge from the extremely limited labeled source data to the unlabeled target data for cross-domain optic disc and cup segmentation. First, we propose to diversify the domain knowledge by a Searching-based Multi-Style Invariant (SMSI) mechanism, enriching the image distributions by creating transformed styles based on the synthesized images through searching strategies. Secondly, considering the similarity of the foreground content within the same category under various styles, a new Class-Prototype Consistency (CPC) mechanism is also introduced. Moreover, we design a Cross-Style Self-supervised Learning (CSSL) strategy with pseudo labels to further boost the overall segmentation of the unlabeled target images. Our proposed method is validated on two cross-domain optic disc and cup segmentation experiments with limited labeled source data available. By outperforming other state-of-the-art UDA segmentation methods, our proposed framework is demonstrated to be more effective and can conquer the domain shift under low resource situations, which is, therefore, more practical and important in real-world applications.

\section{Related Work}
\textbf{Unsupervised domain adaptation} Performing pixel-level domain mapping using image-to-image translation is a typical solution to reduce the domain gap at the appearance level~\cite{isola2017image,zhu2017unpaired}. In addition, feature-level adaptation can also alleviate the cross-domain discrepancy by inducing domain-invariant features learning~\cite{ganin2016domain,tzeng2017adversarial,vu2018advent,liu2020unsupervised,zhang2021deep}. Various methods have involved generative adversarial training, but non-GAN-based techniques have also achieved competitive results, especially those via frequency space learning~\cite{yang2020fda, huang2021,9879210}. These methods mostly conduct frequency alignment or frequency modification to achieve image stylization. By introducing little extra computations to the framework, the frequency learning-based methods can achieve style transformation in a more efficient manner than the GAN-based ones. However, current frequency space methods heavily rely on non-learnable parameter selection, such as parameter $\beta$ in ~\cite{yang2020fda} and parameter $p$ in ~\cite{huang2021}. To avoid massive experiments for selecting the appropriate parameters for the improved synthesized images, we design a SMSI module based on AutoML techniques, which have been widely investigated for efficient medical image analysis~\cite{wang2021bix,xiang2022towards,peng2022hypersegnas}.

\section{Methodology}
\subsection{Searching-based Multi-Style Invariant Mechanism (SMSI)}
To alleviate the domain gap at the appearance level as well as enlarge the data-scarce source domain, we propose a Searching-based Multi-Style Invariant Mechanism (SMSI) for the source domain based on Fourier transform~\cite{yang2020fda}. Specifically, each channel of an input image $x$ is firstly transformed into the frequency space $\mathcal{F}(x)$ via: $\mathcal{F}(x) = \sum_{h,w} x(h,w)e^{-j2\pi(\frac{h}{H}m + \frac{w}{W}n)}$, where $j^2 = -1$. Next, this frequency signal can be further decomposed into an amplitude spectrum~${\mathcal{F}^A}$ and a phase spectrum~${\mathcal{F}^P}$, which respectively represent the low-level (e.g., appearance) and high-level (e.g., content) characteristics of each image~\cite{yang2020fda,jiang2022harmofl}. 

By obtaining the amplitude $\mathcal{F}^A_s$ and phase $\mathcal{F}^P_s$ from each source image $x_{s}$, its corresponding synthesis image in the target-like style can be generated via:
\begin{equation}
    X_{s \rightarrow t} =\mathcal{F}^{-1}[(M_{\beta} \overline {\mathcal{F}^A_t} +(1-M_{\beta})\mathcal{F}^A_s, \mathcal{F}^P_s],
\label{equation-fda}
\end{equation}
where the $\mathcal{F}^{-1}$ is the inverse Fourier transform. To ensure that each synthesis image contains comprehensive appearance-level information under the target distributions, we propose to replace the $\mathcal{F}^A_{s}$ with the average amplitude spectrum from all the target images, denote as $\overline {\mathcal{F}^A_t}$. $M_{\beta}$ is used to control the proportion of the target amplitude during synthesis and controlled by a parameter $\beta \in (0,1)$ , defined as $M_{\beta} = \mathcal{I}_{(h,w)\in [-\beta H : \beta H, -\beta W : \beta W]}$. As indicated in previous works~\cite{yang2020fda},  different $M_{\beta}$ choices can induce distinct domain adaption performance. However, it is cost-intensive to conduct massive experiments for selecting the appropriate parameters for each specific application scenario. To tackle this issue, we propose an efficient searching strategy to find the optimal parameters for the synthesis images which can achieve better cross-domain segmentation performance.  

\begin{figure}
\centering
\includegraphics[width=0.5\textwidth]{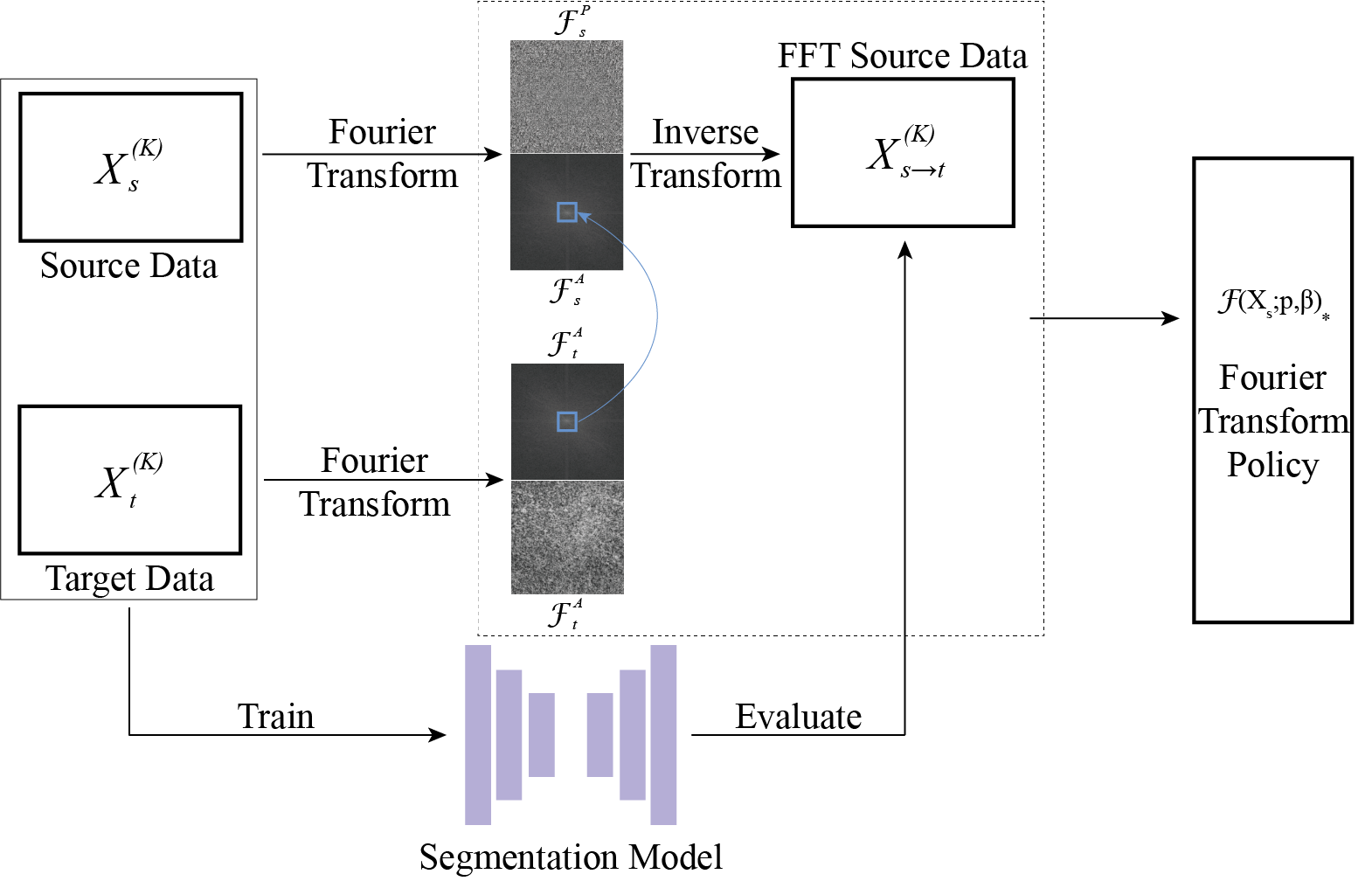}\\
\caption{K-folds automatic Fourier transform searching algorithm. Source dataset $X_s$ and target dataset $X_t$ are split in to k-folds, Fourier transform with different proportion parameter $\beta$ applied to the source dataset, segmentation network trained with target dataset is used to evaluate the transformed data $X_{s \rightarrow t}$. The parameters selected from each fold are appended to the final Fourier transform policy list $\mathcal{F_*}$. }
\label{fig:fda}
\end{figure}

Specifically, the search space is first defined as $\mathcal{F}(X_s;\beta)$. Given the above search space, the search processes are formulated as: (i) Train a plain segmentation model with original source data $X_s$. (ii) Initialized by the model from step (i), several models are further optimized on K groups of synthesis images according to Equation~\ref{equation-fda}. In each group, the $\beta$ for $M_{\beta}$ is randomly initialized within $(0, 1)$. (iii) Let each model in step (ii) learn the ideal $\beta$ by searching controllers and appending to the final policy set following~\cite{47890, lim2019fast}. As the Fourier transformation is only changing the style of each image, instead of its content, the segmentation ground truth of the synthesis images is similar to the original one~\cite{yang2020fda}. Therefore, the objective function of the policy search is designed to maximize the validation dice on transformed data $X_{s \rightarrow t}$ with original source label:
\begin{equation}
    \mathcal{F_*} = argmax\:\mathcal{D}(\theta_G|\mathcal{F}(X_s)),
\end{equation}
where $\theta_G$ is the parameter of the segmentation network used to optimize $\mathcal{L}_{seg}$ and $\mathcal{D}$ is the validation dice. The search controller can be implemented efficiently using the Tree-structured Parzen Estimators algorithm in~\cite{pmlr-v28-bergstra13}. Figure~\ref{fig:fda} indicates the detailed process of the SMSI mechanism. After optimal parameters are determined, each source image can generate $n \times k$ synthesis image by varying the $n$ number of $\beta$ parameters that control the likelihood of $s\rightarrow t$ and taking $k$ average amplitude values from different parts of the target images. This procedure generally expands the source domain dataset and provides a useful regularization technique to increase the diversity of the dataset.

%-------------------------------------------------------------------------

%-------------------------------------------------------------------------

Although training the models with the synthesis images can alleviate the domain gap at the appearance levels, it can still incur domain shifts at the feature level~\cite{hoffman2018cycada,liu2020pdam}. As such, we introduce feature invariant induction learning based on the searching-based multi-style synthesis process. Specifically, additional adversarial domain discriminators are utilized to generate domain-invariant features for the synthesis images and target images on top of the traditional supervised loss. Denote that the source dataset is $X_s \subset R^{H \times W \times 3}$ with ground truth C-class segmentation maps $Y_s \subset (1,C)^{H \times W}$, synthesis source dataset is  $X_{s\rightarrow t}$ with the same ground truth maps. The target dataset is $X_t$ with no ground truth label. A discriminator $\theta_D$ is trained adversarially to distinguish between the synthesis source set and target set with discrimination loss $\mathcal{L}_{D}$. Simultaneously, the segmentation network is trained to fool the discriminator as:
\begin{equation}
    \min_{\theta_D}\frac{1}{|X_{s\rightarrow t}|} \sum_{x_{s\rightarrow t}}\mathcal{L}_{D}(I_{x_{s\rightarrow t}},1) + \frac{1}{|X_t|} \sum_{x_t}\mathcal{L}_{D}(I_{x_t},0) \qquad \min_{\theta_G}\frac{1}{|X_t|} \sum_{x_t}\mathcal{L}_{D}(I_{x_t},1)
\end{equation}
where $I_{x_{s\rightarrow t}}$ and $I_{x_t}$ are the weighted self-information maps following ADVENT~\cite{vu2018advent}. Summarily, there are two discriminators $D_1$ and $D_2$ implemented to distinguish (i) $X_{s\rightarrow t}$ and $X_t$ (ii) $X_s$ and $X_t$ as indicated in the top part of Figure~\ref{fig:overview}.

\subsection{Class-Prototype Consistency Mechanism (CPC)}
The synthesis images produced from the SMSI mechanism and their corresponding source images should have the same image content but in different styles. Motivated by previous works that the class-aware features under the same category should maintain the same across different domains~\cite{zheng2020cross}, we propose a class-prototype consistency mechanism for the synthesis images. The class prototype is created by using the high-level feature maps from the model's encoder and the ground truth source masks. The source masks are first resized and converted into binary masks for each class. Then, they are multiplied by the feature maps extracted from synthesis images and source images respectively, generating class-relevant masked feature maps. Taking a global average pooling further converts the feature maps into class prototype vectors. Global average pooling has the ability to sum out the spatial data and enforce the correspondences between feature maps and classes. Denote the mask of class $c$ as $M^{c}$ and $f_s$,  $f_{s\rightarrow t}$ as the feature maps, the class prototypes of $c$ class are defined as:
\begin{equation}
    p^{c}_{s} = \frac{1}{N_{h\times w}}\sum_{c} M^{c} f_s  \qquad p^{c}_{s\rightarrow t} = \frac{1}{N_{h\times w}}\sum_{c} M^{c} f_{s\rightarrow t}
\end{equation}
where $h$ and $w$ are the height and width of the feature maps. This masked pooling technique enables the network to focus on the target content of images instead of intensity and illumination variation. To narrow the gap between the features under the same class in different synthesis domains, we propose to enlarge the similarity between them. The overall CPC Mechanism is illustrated in the bottom part of Figure~\ref{fig:overview}, where binary masks for optic cups and discs are used. Specifically, the overall consistency loss function can be defined as:
\begin{equation}
    \mathcal{L}_{con} = \sum_{c=(0,1)} \Big(1 - \frac{p^{c}_{s} \cdot p^{c}_{s\rightarrow t}}{max(||p^{c}_{s}||_2 \cdot ||p^{c}_{s\rightarrow t}||_2,\epsilon)}\Big),
\end{equation}
where $\epsilon$ is the small value to avoid division by zero, $p^{c}_{s}$ and $p^{c}_{s\rightarrow t}$ are the prototypes of the class $c$ for the features from the FFT synthesized images and source images. The overall optimization function for the segmentation network with SMSI and CPC mechanism is defined as:
\begin{equation}
        \mathcal{L}_{total} = \mathcal{L}_{seg}(X_s,Y_{s})+ \mathcal{L}_{seg}(X_{s\rightarrow t},Y_{s}) + \lambda(\mathcal{L}_{D_1}+\mathcal{L}_{D_2}) + \mathcal{L}_{con}.
\label{equation-total}
\end{equation}

\subsection{Cross-Style Self-supervised Learning (CSSL)~\label{sec-pl}}
When dealing with model adaptation towards the target domain, a consummate resource is the target ground truth masks, which are not available in UDA settings. As compensation, highly-confident pseudo labels can be created for unlabeled target images by using prediction probabilities $P^v$ from the trained model $\theta_G^1$ on the v-th pixel. The pseudo labels can be defined as $\hat{y}^v_t = I[P^v \geq \gamma]$, where $I$ is the indicator function and $\gamma \in (0,1)$ is the probability threshold to determine the binary mask.
%-------------------------------------------------------------------------
\begin{figure}
\centering
\includegraphics[width=0.5\textwidth]{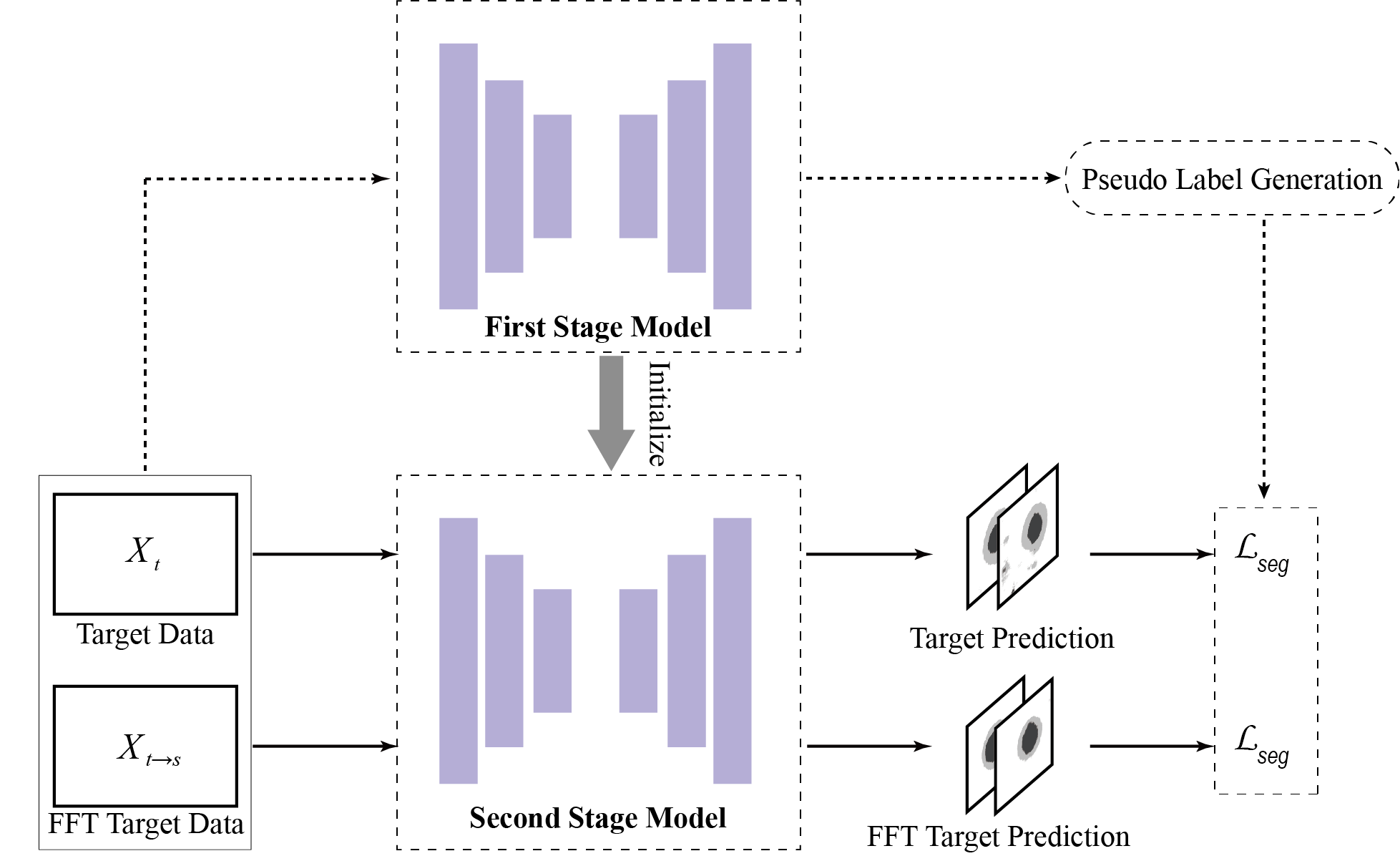}\\
\caption{Demonstration of the pseudo label self-supervised learning process. Specifically, a second stage model $\theta_G^2$ is initialized by $\theta_G^1$ and trained over the target data $X_t$ and the FFT target data $X_{t\rightarrow s}$ with pseudo labels. }
\label{fig:plabel}
\end{figure}
%-------------------------------------------------------------------------
However, solely training the model with the target pseudo labels brings noise to the optimization process due to the gap between the pseudo and real labels. To stabilize the training process, we propose a cross-style self-supervised learning strategy, to jointly re-train the model using the target images $X_t$ and the Fourier transformed target images with source-like styles $X_{t\rightarrow s}$ and their pseudo labels $\hat{y}^v_t$, and $\hat{y}^v_{t\rightarrow s}$, respectively. The source-like synthesized images are obtained following the process in Section 2.1. Since the segmentation learning for the model in the first stage is based on the annotated source data, the pseudo labels for the source-like synthesis images contain less noise and therefore can be a complement to the target supervised loss. Figure~\ref{fig:plabel} demonstrates these two supervised segmentation losses. In general, the model learns combined distribution and gets further improvement using self-supervised training in the second stage.

\section{Experiments}
\subsection{Datasets and Implementation Details}
The experiments aim to segment the cup and disc components in the multi-center fundus images, which are obtained from various patients using different eye examination equipment. The source dataset REFUGE~\cite{orlando2020refuge} contains 400 annotated images. There are two target datasets, RIM ONE-r3~\cite{fumero2015interactive} contains 99 training images and 60 testing images, and Drishti-GS~\cite{sivaswamy2014drishti} contains 50 training images and 51 testing images. All datasets used are publicly available. In our experiments with few labeled source data, only 10 random source images are accessible during model training. For the image synthesis process in the first stage, $n=3$ and $k=5$ are selected for each source image, with a total of 150 FFT source images generated. In the second stage, $n=3$ and $k=1$ are selected for each target image, 150 and 297 FFT target images are generated respectively. The experiment results under other selections for the source images as well as more detailed experimental and implementation settings are available in the supplementary material. 

% For Fourier transform, the optimal $\beta$ parameters searched are indicated in table~\ref{table0.1}. 

% \begin{table}[!htbp]
% \caption{Selection of the parameter from automatic FFT}
% \centering
% \begin{tabular}{c|ccc}
% \hline
% Fourier Transform                     & \multicolumn{3}{c}{$\beta$ parameter}                                 \\ \hline
% REFUGE to RIM ONE-r3 & \multicolumn{1}{c|}{0.2702} & \multicolumn{1}{c|}{0.0952} & 0.2390 \\
% RIM ONE-r3 to REFUGE & \multicolumn{1}{c|}{0.1503} & \multicolumn{1}{c|}{0.1766} & 0.0176 \\
% REFUGE to Drishti-GS & \multicolumn{1}{c|}{0.2067}       & \multicolumn{1}{c|}{0.3439}       &     0.7790   \\
% Drishti-GS to REFUGE & \multicolumn{1}{c|}{0.5241} & \multicolumn{1}{c|}{0.0758} & 0.5100 \\ \hline
% \end{tabular}
% \label{table0.1}
% \end{table}

The network used in our experiments is a MobileNetv2 with a DeepLabv3+ backbone based on the structure in \cite{wang2019boundary}. A semantic segmentation module called Atrous Spatial Pyramid Pooling (ASPP) in DeepLab re-samples a given feature layer at various rates before convolution. The overall model size is 7.62M and the inference time is 30.57s for one image. In the first stage, the segmentation model is trained with Adam optimizer under a 1e-3 learning rate, the discriminators have been trained with SGD optimizer with a 2.5e-5 learning rate, 8 batch size, and 200 training epochs. The weighting factor $\lambda$ in Equation~\ref{equation-total} is set as $0.5$. In the second stage, the segmentation model is trained with Adam optimizer with a 2e-3 learning rate, 8 batch size, and 20 training epochs. The probability threshold $\gamma = 0.75$ is used to generate the pseudo labels. Segmentation results are evaluated by the Dice coefficient and Average Surface Distance (ASD). The framework is implemented on Pytorch 1.7.1 using a NVIDIA RTX3090 GPU.

\subsection{Comparison Experiments}
The proposed method is compared to state-of-the-art (SOTA) unsupervised domain adaptation methods, as well as two recent UDA approaches particularly for few labeled source images. In supplementary material, extra SOTA fundus image segmentation methods~\cite{Wang2019,Feng2022,liu2022cada,zhang2021tau,liu2022ecsd} are also compared. CyCADA~\cite{hoffman2018cycada} translates the source images into the target style using cycle-consistent adversarial networks and trains the adversarial network with the translated images. AdvEnt~\cite{vu2018advent} brings in entropy loss and adversarial loss respectively to address the domain shift problem. FDA~\cite{yang2020fda} adopts frequency swap method for image stylization and evaluate the segmentation model with multi-band transfer. PixMatch~\cite{melas2021pixmatch} develops a new component to ensure that the model's predictions on a target image and a perturbed version of the same image are pixel-wise consistent. LTIR~\cite{kim2020learning} learns texture invariant features from different domains using Style-Wrap to change the images' appearance. Consider from another perspective, the two recently-developed methods MT~\cite{zhao2021mt} and PCS~\cite{yue2021prototypical} have a similar experimental setting, they both focus on domain adaptation with few source data. MT follows the mean teacher paradigm and adopts dual teacher models to provide both semantic and structural knowledge to the student model, whereas PCS performs in-domain and cross-domain learning using prototypes from feature memory banks.  Some other latest fundus image segmentation baselines are also evaluated. BEAL~\cite{wang2019boundary} suggests boundary prediction and entropy-driven during adversarial training and achieves excellent results for cross-domain prediction. DPL~\cite{chen2021source} is a novel proposal for source-free domain adaptation in the field of fundus image segmentation, with a pseudo-label denoising technique. It utilizes a pre-trained source model to generate pseudo-labels. For comparison, we follow the same experimental settings in these segmentation models, i.e., only 10 randomly selected source images will be accessible throughout the experiment, even for the pre-trained model. 

{\bf Quantitative analysis.} As indicated in Table \ref{table:results}, the segmentation performance of all other comparison UDA approaches is at the same level. This indicates their adaptation abilities are limited due to the lack of sufficient supervision learning. For the MT~\cite{zhao2021mt} and PCS~\cite{yue2021prototypical} which were originally designed for UDA with few labeled source data, we notice that their performance is suboptimal. For MT, the lack of consideration of the cross-domain category information makes the model learn insufficient semantic-level knowledge given the extremely limited labeled source data for segmentation supervision learning, which further incurs inferior performance on the target testing data. Although PCS proposes a class-aware UDA framework, it was particularly designed for UDA classification under the small domain gap. When validated on the UDA fundus image segmentation with a large domain bias, its segmentation results are limited by ignoring the appearance-level domain bias and the particular designs for segmentation. On the other hand, our method can tackle the aforementioned challenges by the SMSI for appearance-level adaption, CPC for cross-domain category-aware information processing, and the CSSL for further performance gain without auxiliary annotations. Overall, our method has outperformed others, achieving 6.22\% Dice, 10.71 ASD pixel for RIM-ONE- r3 and 4.44\% Dice, 5.92 ASD pixel for Drishti-Gs. We have also conducted a two-tailed paired t-test on the comparison studies, and given the p-value smaller than 0.01, our improvements are statistically significant.

\begin{table}
\centering
\resizebox{0.97\linewidth}{!}{%
\begin{tabular}{c|c|c|c|c|c|c|cl} 
\cline{1-8}
\multicolumn{2}{c|}{\multirow{2}{*}{Methods}}                                                                     & \multicolumn{3}{c|}{Dice Metric~[\%]~$\uparrow$}            & \multicolumn{3}{c}{ASD Metric [pixel]~$\downarrow$}           & \multicolumn{1}{c}{}  \\ 
\cline{3-8}
\multicolumn{2}{l|}{}                                                                                             & Cup            & Disc           & Average        & Cup            & Disc           & Average        &                       \\ 
\cline{1-8}
\multicolumn{1}{l}{}                                                                              & \multicolumn{7}{c}{\textbf{RIM-ONE-r3}}                                                                             & \multicolumn{1}{c}{}  \\ 
\cline{1-8}
& CyCADA~\cite{hoffman2018cycada}     & 69.94 (0.58)         & 72.50 (0.37)        & 71.32 (0.48)        & 19.55 (0.22)       & 37.00  (0.35)        & 28.28 (0.29)       &                       \\
                                                                                                  & AdvEnt~\cite{vu2018advent}       & 67.73 (0.63)         & 78.54 (1.19)         & 73.34  (0.91)        & 30.96  (1.79)        & 32.48  (1.55)        & 31.72  (1.67)        &                       \\
                                                                                  &
                                     FDA~\cite{yang2020fda}       &  
                                     69.38 (0.20) &    78.07  (1.10)    &    73.72 (0.65)      &   21.15 (1.47)      &    28.86  (1.41)     &  25.01  (1.44)        &                       \\
                                                                                                  & PixMatch~\cite{melas2021pixmatch}     & 64.91 (1.14)          & 75.88 (1.93)         & 70.39   (1.53)       & 18.60  (0.39)        & 30.50 (1.42)      & 24.55  (0.91)          &                       \\
                                                                                                  & LTIR~~\cite{kim2020learning}         & 65.84  (0.09)        & 78.01  (1.97)        & 71.92  (1.03)        & 24.72  (0.98)        & 29.51   (1.46)       & 27.11   (1.22)       &                       \\ 
% \cline{1-8}
             & MT~\cite{zhao2021mt}           & 63.50   (1.51)       & 67.92  (1.05)       & 65.71  (1.28)       & 20.72   (1.88)      & 39.26  (1.86)        & 29.99    (1.87)      &                       \\
                                                                                                  & PCS~\cite{yue2021prototypical}          & 60.77  (1.19)        & 73.90  (1.50)         & 67.33    (1.35)      & 24.20  (1.86)        & 32.13  (1.51)        & 28.16    (1.69)      &                       \\ 
\cline{1-8}                                                                                                                                                                           & BEAL~\cite{wang2019boundary}         & 67.69 (1.49)         & 78.88  (1.23)        & 73.29    (1.36)      & 21.36 (1.93)         & 34.95    (1.79)      & 28.16  (1.86)        &                       \\
                                                                                                  & DPL~\cite{chen2021source}          & 68.58 (0.33)         & 87.61  (0.61)      & 78.02 (0.47)         & 12.46 (0.59)          & 19.01  (0.45)        & 15.74 (0.52)         &                       \\                                                                
% \cline{1-8}
% \multirow{2}{*}{all labeled source}                                                                                 & CADA~\cite{liu2022cada}        & 64.04          & 76.64         & 70.34          & -          & -         & -         &                       \\
%                                                                                                   & ECSD-Net~\cite{liu2022ecsd}         & 80.20         & 86.50          & 83.35          & -          & -        & -          &                       \\
                                                                                      & \textbf{Ours} & \textbf{78.16 (0.96)} & \textbf{88.45 (0.36)} & \textbf{83.30 (0.66)} & \textbf{9.82 (0.65)} & \textbf{11.78 (0.75)} & \textbf{10.80 (0.70)} &                       \\ 
\cline{1-8}
\multicolumn{1}{l}{}                                                                              & \multicolumn{7}{c}{\textbf{Drishti-GS}}                                                                             & \multicolumn{1}{c}{}  \\ 
\cline{1-8}
& CyCADA~\cite{hoffman2018cycada}    & 78.85 (0.80)        & 92.15  (0.52)        & 85.50   (0.66)      & 15.33   (0.30)       & 12.84   (1.85)       & 14.09  (1.08)        &                       \\
                                                                                                  & AdvEnt~\cite{vu2018advent}       & 79.17 (0.79)         & 91.47 (0.47)         & 85.32  (0.63)        & 15.42  (0.85)        & 15.08 (1.29)         & 15.25    (1.07)      &                       \\   &
                                     FDA~\cite{yang2020fda}          & 
                                     83.57 (0.40) &   94.13  (0.67)      &    88.85  (0.54)    &     12.40 (0.25)    &   7.68  (0.50)       &   10.04  (0.38)      &                       \\
                                                                                                  & PixMatch~\cite{melas2021pixmatch}     & 76.45  (1.60)        & 91.97 (0.47)          & 84.21   (1.04)       & 17.06  (1.13)         & 10.92  (1.04)        & 13.99  (1.09)        &                       \\
                                                                                                  & LTIR~~\cite{kim2020learning}         & 80.82   (1.00)       & 92.86   (0.73)       & 86.84   (0.87)       & 13.29    (0.74)      & 8.83  (0.94)         & 11.06  (0.84)        &                       \\ 
% \cline{1-8}
               & MT~\cite{zhao2021mt}           & 72.82  (1.54)        & 90.37  (0.99)        & 81.59     (1.27)     & 19.43   (0.30)       & 13.11   (1.73)       & 16.39  (1.02)        &                       \\
                                                                                                  & PCS~\cite{yue2021prototypical}          & 74.88  (1.19)        & 88.33 (0.03)         & 81.21   (0.61)       & 19.81  (1.45)        & 18.90  (0.78)        & 19.56  (1.12)        &                       \\ 
\cline{1-8}
% \multirow{2}{*}{all labeled source}                                                                                 & CADA~\cite{liu2022cada}        & 84.00        & 89.00        & 86.50        & -          & -         & -         &                       \\
%                                                                                                   & ECSD-Net~\cite{liu2022ecsd}        & 87.60        & 96.50          & 92.05          & -          & -          &-          &                       \\
%  \cline{1-8}  
                                                                             & BEAL~\cite{wang2019boundary}         & 75.91 (1.29)      & 93.44 (0.36)      & 84.68  (0.83)      & 17.48  (1.47)       & 10.42  (0.36)        & 13.95  (0.92)        &                       \\
                                                                                                  & DPL~\cite{chen2021source}          & 78.60 (0.17)        & 95.28  (0.82)        & 86.94 (0.50)         & 19.99  (0.82)        & 6.07  (0.16)         & 13.03 (0.49)         &                       \\
                                                                                                  & \textbf{Ours} & \textbf{83.64(0.20)} & \textbf{95.47(0.23)} & \textbf{89.56(0.22)} & \textbf{11.04(0.51)} & \textbf{5.25(0.25)}  & \textbf{8.14(0.38)}  &                       \\
\cline{1-8}
\end{tabular}}
\caption{Experimental results for comparison of different domain adaptation approaches in terms of Dice and ASD metrics on RIM-ONE-r3 and Drishti-GS target datasets. The numbers shown in the table are average values, taken from three sets of experiments under different groups of randomly selected 10 labeled source images. The numbers in parenthesis are the SD values.}
\label{table:results}
\end{table}

\begin{figure}
\centering
\includegraphics[width=\textwidth]{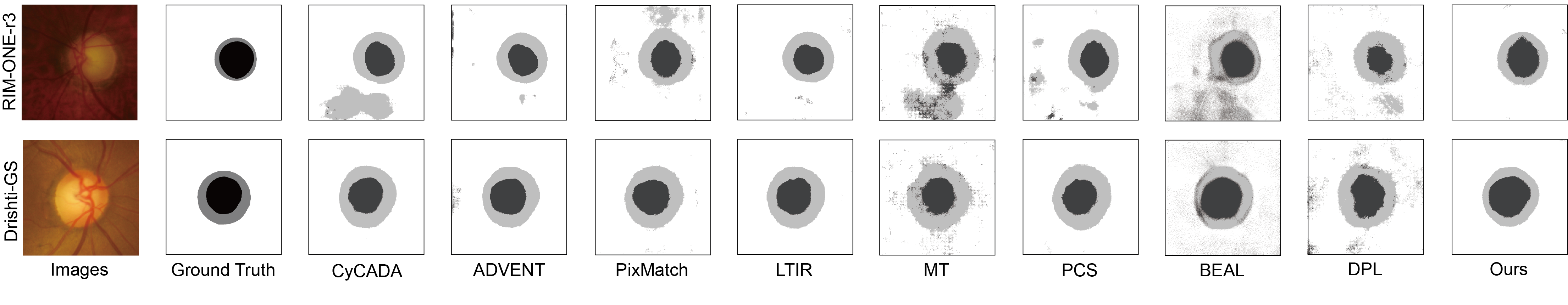}
\caption{Segmentation results from some of the comparison experiments.} 
\label{fig:results}
\end{figure}

{\bf Qualitative analysis.} As presented in Figure \ref{fig:results}, the segmentation results of some experiments show that focusing on content rather than appearance enables the network to better distinguish target objects from irrelevant backgrounds. The segmentation predictions from several comparison methods are significantly distracted by the background noise. Additionally, regardless of domain differences, the network faces difficulties when attempting to determine the spatial prior knowledge of the optic disc and optic cup. Our predictions alleviate these issues, have a much clear boundary between the cup and disc, and exhibit much fewer background segmentation error. 

% Pseudo label learning with $X_t$ and $X_{t\rightarrow s}$, Plain Pseudo label learning with $X_t$

\subsection{Ablation Studies}
Ablation studies are conducted to evaluate the effectiveness of our proposed modules. In Table \ref{table:ablation}, the source-only and target-only experiments provide lower and upper bounds of this setting. The source-only experiment trains the segmentation network using only source images and directly adapts to the target domain. By contrast, a target-only experiment trains the network using annotated target data under a supervised learning setting. The first implemented module is standard adversarial training with additional discriminator loss. The improvement under two metrics suggests that the concept of adversarial training can be drawn on this task. Then SMSI further boosts the performance by diversifying the source styles and inducing the domain invariant feature generation, providing a large quantity of labeled data for few-shot learning. In addition, a novel class prototype consistency loss allows the network to particularly align the features at the category level. Both Dice and ASD metrics indicate that the proposed method significantly increases the adaptation ability of the segmentation network with limited labeled data. 

\begin{table}[!htbp]
\centering
\resizebox{0.99\linewidth}{!}{%
\begin{tabular}{ccccccc}
\hline
\multicolumn{1}{c|}{Settings}                       & \multicolumn{3}{c|}{Dice Metric {[}\%{]}~$\uparrow$}                                                                       & \multicolumn{3}{c}{ASD Metric {[}pixel{]}~$\downarrow$}                                                 \\ \cline{2-7} 
\multicolumn{1}{c|}{}                                 & \multicolumn{1}{c|}{Cup}            & \multicolumn{1}{c|}{Disc}           & \multicolumn{1}{c|}{Average}        & \multicolumn{1}{c|}{Cup}            & \multicolumn{1}{c|}{Disc}           & Average        \\ \hline
\multicolumn{7}{c}{\textbf{RIM-ONE-r3}}                                                                                                                                                                                                                              \\ \hline
\multicolumn{1}{c|}{Source only}                      & \multicolumn{1}{c|}{58.75}          & \multicolumn{1}{c|}{64.84}          & \multicolumn{1}{c|}{61.79}          & \multicolumn{1}{c|}{26.92}          & \multicolumn{1}{c|}{43.10}          & 35.01          \\
\multicolumn{1}{c|}{Adversarial baseline with $X_s$ and $X_t$}             & \multicolumn{1}{c|}{66.95}          & \multicolumn{1}{c|}{69.46}          & \multicolumn{1}{c|}{68.20}          & \multicolumn{1}{c|}{26.65}          & \multicolumn{1}{c|}{46.94}          & 36.80          \\
\multicolumn{1}{c|}{+ SMSI with $X_s$ and $X_{s\rightarrow t}$} & \multicolumn{1}{c|}{67.37}          & \multicolumn{1}{c|}{77.61}          & \multicolumn{1}{c|}{72.49}          & \multicolumn{1}{c|}{21.07}          & \multicolumn{1}{c|}{27.72}          & 24.40          \\
\multicolumn{1}{c|}{+ Class-prototype consistency}    & \multicolumn{1}{c|}{68.93}          & \multicolumn{1}{c|}{82.52}          & \multicolumn{1}{c|}{75.72}          & \multicolumn{1}{c|}{21.81}          & \multicolumn{1}{c|}{19.86}          & 20.84          \\
\multicolumn{1}{c|}{+ Plain pseudo label learning with $X_t$ }    & \multicolumn{1}{c|}{77.03}          & \multicolumn{1}{c|}{85.79}          & \multicolumn{1}{c|}{81.41}          & \multicolumn{1}{c|}{10.13}          & \multicolumn{1}{c|}{14.74}          & 12.43          \\
\multicolumn{1}{c|}{+ CSSL with $X_t$ and $X_{t\rightarrow s}$}          & \multicolumn{1}{c|}{78.16} & \multicolumn{1}{c|}{88.45} & \multicolumn{1}{c|}{83.30} & \multicolumn{1}{c|}{9.82} & \multicolumn{1}{c|}{11.78} & 10.80 \\
\multicolumn{1}{c|}{Target only}                      & \multicolumn{1}{c|}{80.88}          & \multicolumn{1}{c|}{95.57}          & \multicolumn{1}{c|}{88.22}          & \multicolumn{1}{c|}{10.21}          & \multicolumn{1}{c|}{5.20}           & 7.71           \\ \hline
\multicolumn{7}{c}{\textbf{Drishti-GS}}                                                                                                                                                                                                                              \\ \hline
\multicolumn{1}{c|}{Source only}                      & \multicolumn{1}{c|}{76.55}          & \multicolumn{1}{c|}{84.53}          & \multicolumn{1}{c|}{80.54}          & \multicolumn{1}{c|}{18.77}          & \multicolumn{1}{c|}{23.08}          & 20.93          \\
\multicolumn{1}{c|}{Adversarial baseline with $X_s$ and $X_t$}             & \multicolumn{1}{c|}{76.86}          & \multicolumn{1}{c|}{93.72}          & \multicolumn{1}{c|}{85.29}          & \multicolumn{1}{c|}{16.68}          & \multicolumn{1}{c|}{9.68}           & 13.18          \\
\multicolumn{1}{c|}{+ SMSI with $X_s$ and $X_{s\rightarrow t}$} & \multicolumn{1}{c|}{83.45}          & \multicolumn{1}{c|}{93.12}          & \multicolumn{1}{c|}{88.28}          & \multicolumn{1}{c|}{12.37}          & \multicolumn{1}{c|}{8.60}           & 10.48          \\
\multicolumn{1}{c|}{+ Class-prototype consistency}    & \multicolumn{1}{c|}{84.76}          & \multicolumn{1}{c|}{94.26}          & \multicolumn{1}{c|}{89.51}          & \multicolumn{1}{c|}{11.02}          & \multicolumn{1}{c|}{7.04}           & 9.03           \\
\multicolumn{1}{c|}{+ Plain pseudo label learning  with $X_t$}    & \multicolumn{1}{c|}{84.24}          & \multicolumn{1}{c|}{94.88}          & \multicolumn{1}{c|}{89.56}          & \multicolumn{1}{c|}{10.52}          & \multicolumn{1}{c|}{5.93}          & 8.23          \\
\multicolumn{1}{c|}{+ CSSL with $X_t$ and $X_{t\rightarrow s}$}          & \multicolumn{1}{c|}{83.64} & \multicolumn{1}{c|}{95.47} & \multicolumn{1}{c|}{89.56} & \multicolumn{1}{c|}{11.04} & \multicolumn{1}{c|}{5.25}  & 8.14  \\
\multicolumn{1}{c|}{Target only}                      & \multicolumn{1}{c|}{84.19}          & \multicolumn{1}{c|}{97.07}          & \multicolumn{1}{c|}{90.63}          & \multicolumn{1}{c|}{11.11}          & \multicolumn{1}{c|}{3.85}           & 7.48           \\ \hline
\end{tabular}}
\caption{Ablation results with implemented domain adaptation modules in terms of Dice and ASD metrics on RIM-ONE-r3 and Drishti-GS target datasets.}
\label{table:ablation}
\end{table}

\begin{figure}
\centering
\includegraphics[width=0.8\textwidth]{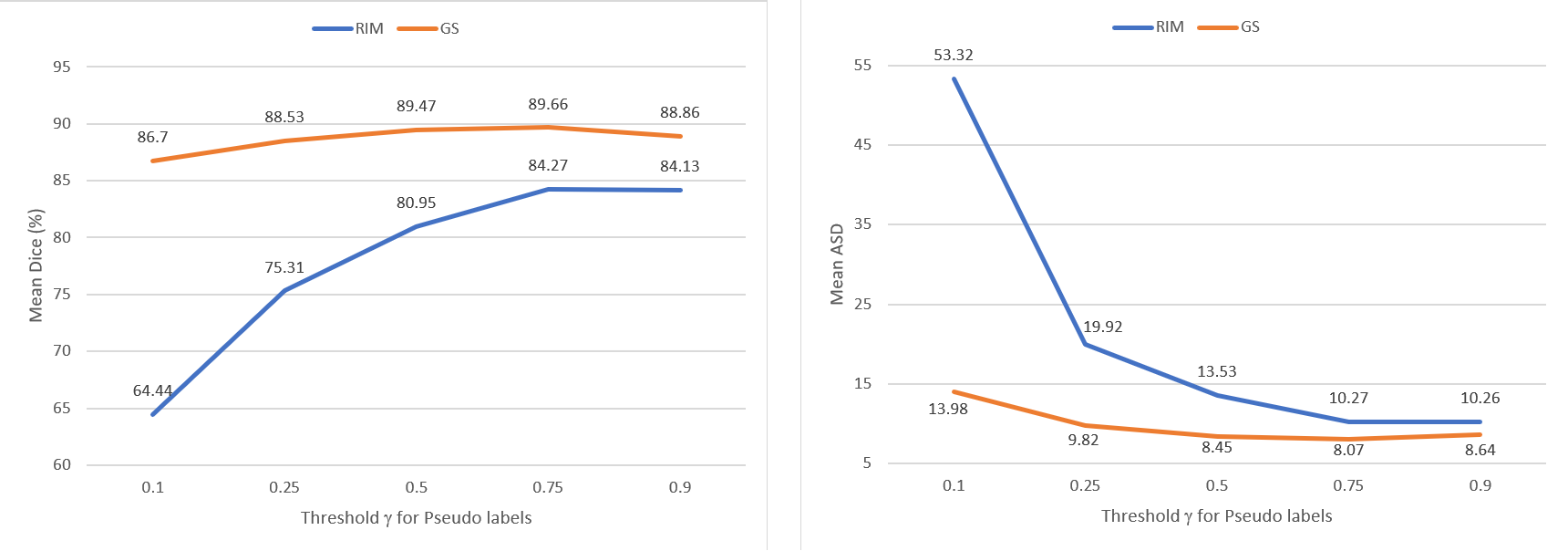}
\caption{UDA Segmentation performance under different selections of the threshold $\gamma$ for the pseudo label learning stage. RIM and GS indicate our experiment settings of using the RIM-ONE-r3 and Drishti-GS as the target domain, respectively.}
\label{fig-pl-thres}
\end{figure}

On top of these modules at the first stage, our proposed Cross-Style Self-supervised Learning (CSSL) module brings an improvement of about 14\% in dice value and 18 pixels in ASD over the non-adaptation model. We also conduct ablation experiments by conducting the self-supervised learning only on the target images, which introduces less performance gain than the CSSL. This further demonstrates the claim in Section~\ref{sec-pl} that our CSSL can alleviate the noises from the pseudo labels and lead to a better self-supervised segmentation performance. By jointly conducting our proposed strategies, the segmentation performance of the source-only model can be improved to a level similar to that of the fully supervised model. In addition, we also explore the models' effectiveness under different thresholds $\gamma$ for the pseudo label learning stage introduced in Section~\ref{sec-pl}. As shown in Figure~\ref{fig-pl-thres}, the best segmentation performance under the Dice and ASD metrics is obtained under both settings when the threshold $\gamma$ is $0.75$.   

\section{Conclusion}
In this work, we propose a novel framework for domain adaptive optic disc and cup segmentation given only a few labeled source data. To alleviate the domain bias issue under the data-scarce setting, the SMSI, CCP, and CSSL modules are designed. In comparison to alternative domain adaptation strategies and even fully supervised networks, the model has been trained to reach competitive outcomes. In this work, we notice the principle bias between the two domains results from the different image styles due to the device variation, and there are no severe distinctions between the morphological structures for the foreground objects in the two domains. As such, future studies are suggested on the cross-domain segmentation problems with larger distinctions in the labeling space.

\section{Supplementary Material}
\begin{table}[!htbp]
\centering
\begin{tabular}{c|ccc}
\hline
Fourier Transform                     & \multicolumn{3}{c}{$\beta$ parameter}                                 \\ \hline
REFUGE to RIM ONE-r3 & \multicolumn{1}{c|}{0.7963} & \multicolumn{1}{c|}{0.8884} & 0.6185 \\
RIM ONE-r3 to REFUGE & \multicolumn{1}{c|}{0.1112} & \multicolumn{1}{c|}{0.3517} & 0.9209 \\
REFUGE to Drishti-GS & \multicolumn{1}{c|}{0.4235}       & \multicolumn{1}{c|}{0.0720}       &     0.2372   \\
Drishti-GS to REFUGE & \multicolumn{1}{c|}{0.6194} & \multicolumn{1}{c|}{0.5651} & 0.0056 \\ \hline
\end{tabular}
\caption{The selection of the optimal $\beta$ parameters for Fourier transform. Referring to the searching-based multi-style invariant mechanism (SMSI), the transformation of REFUGE to the other two target domains is required in the first stage as $X_{s \rightarrow t }$, transformation of two target domains to REFUGE is required in the second stage as $X_{t \rightarrow s}$.
}
\label{table0.1}
\end{table}

\begin{figure}
\centering
\includegraphics[width=0.9\textwidth]{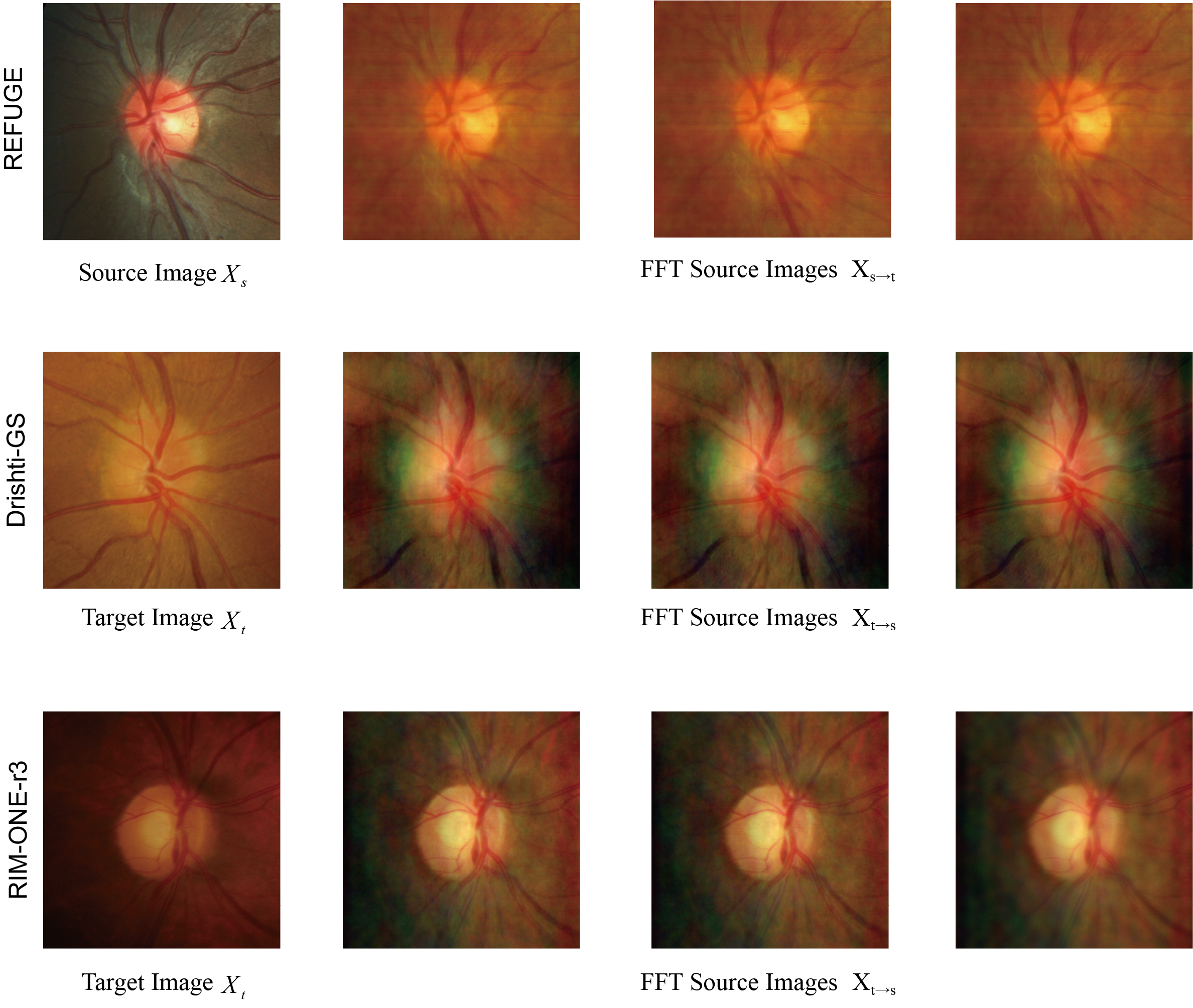}
\caption{Exhibition of synthesis images based on SMSI. Referring to Table~\ref{table0.1} for each dataset, three optimal $\beta$ values are selected.} 
\label{fig:synthesis2}
\end{figure}

\begin{figure}
\centering
\includegraphics[width=\textwidth]{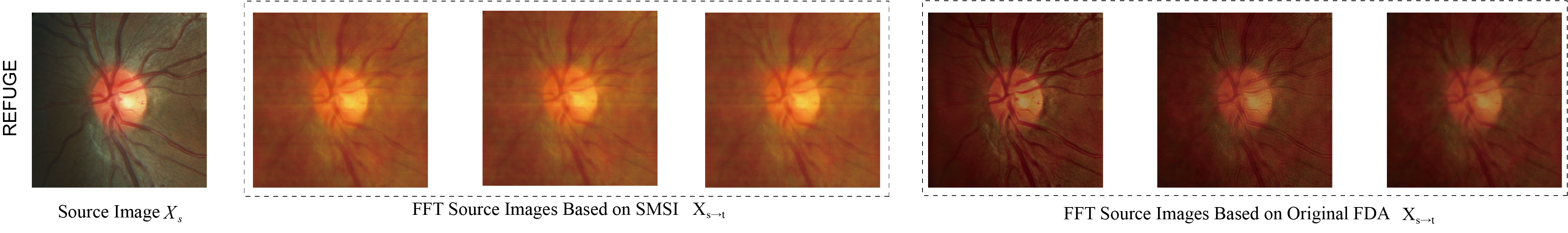}
\caption{Exhibition of synthesis images based on original FDA and SMSI. For original FDA, $\beta = 0.01/0.05/0.09$ values are selected. } 
\label{fig:synthesis1}
\end{figure}

% \subsection*{Experiment results under other selections for the source images}

\begin{table}[!htbp]
\centering
\begin{tabular}{cccccccl}
\cline{1-7}
\multicolumn{1}{c|}{Methods}                           & \multicolumn{3}{c|}{Dice Metric {[}\%{]}}                                                                       & \multicolumn{3}{c}{ASD Metric {[}pixel{]}}                                                 &  \\ \cline{2-7}
\multicolumn{1}{c|}{}                                  & \multicolumn{1}{c|}{Cup}            & \multicolumn{1}{c|}{Disc}           & \multicolumn{1}{c|}{Average}        & \multicolumn{1}{c|}{Cup}            & \multicolumn{1}{c|}{Disc}           & Average        &  \\ \cline{1-7}
\multicolumn{7}{c}{\textbf{RIM-ONE-r3}}                                                                                                                                                                                                                               &  \\ \cline{1-7}
\multicolumn{1}{c|}{10 shots}           & \multicolumn{1}{c|}{78.16}          & \multicolumn{1}{c|}{88.45}          & \multicolumn{1}{c|}{83.30}          & \multicolumn{1}{c|}{9.82}          & \multicolumn{1}{c|}{11.78}          & 10.80         &  \\
\multicolumn{1}{c|}{20 shots}           & \multicolumn{1}{c|}{79.45}          & \multicolumn{1}{c|}{88.41}          & \multicolumn{1}{c|}{83.93}          & \multicolumn{1}{c|}{9.27}          & \multicolumn{1}{c|}{10.97}          & 10.12         &  \\
\multicolumn{1}{c|}{30 shots} & \multicolumn{1}{c|}{79.68} & \multicolumn{1}{c|}{88.25} & \multicolumn{1}{c|}{83.97} & \multicolumn{1}{c|}{8.99} & \multicolumn{1}{c|}{10.65} & 9.82 &  \\ \cline{1-7}
\multicolumn{7}{c}{\textbf{Drishti-GS}}                                                                                                                                                                                                                               &  \\ \cline{1-7}
\multicolumn{1}{c|}{10 shots}           & \multicolumn{1}{c|}{83.64}          & \multicolumn{1}{c|}{95.47}          & \multicolumn{1}{c|}{89.56}          & \multicolumn{1}{c|}{11.04}          & \multicolumn{1}{c|}{5.25}          & 8.14         &  \\
\multicolumn{1}{c|}{20 shots}           & \multicolumn{1}{c|}{82.68}          & \multicolumn{1}{c|}{96.49}          & \multicolumn{1}{c|}{89.59}          & \multicolumn{1}{c|}{11.33}          & \multicolumn{1}{c|}{4.02}          & 7.68          &  \\
\multicolumn{1}{c|}{30 shots} & \multicolumn{1}{c|}{84.93} & \multicolumn{1}{c|}{96.27} & \multicolumn{1}{c|}{90.60} & \multicolumn{1}{c|}{10.06} & \multicolumn{1}{c|}{4.25} & 7.15 &  \\ \cline{1-7}
\end{tabular}
\label{table2}
\caption{Experimental results of our proposed method in terms of Dice and ASD metrics on RIM-ONE-r3 and Drishti-GS target datasets with randomly selected 10, 20, and 30 shots source data.}
\end{table}

% s-cuda is for domain adaptation under noisy labels, and I think it cannot be directly compared with.
% The data split of the NENet method is quite different from all the others, so I also remove them.

% The training and testing images in the target domain used by all comparison methods are the same. The results for other methods are directly referenced from the articles.

\begin{table}[!htbp]
\centering
\begin{tabular}{cccc}
\hline
\multicolumn{1}{c|}{Methods}       & \multicolumn{1}{c|}{Cup  Dice}      & \multicolumn{1}{c|}{Disc Dice}      & Average Dice   \\ \hline
\multicolumn{4}{c}{\textbf{RIM-ONE-r3}}                                                                                         \\ \hline
% \multicolumn{1}{c|}{NENet~\cite{pachade2021nenet}}         & \multicolumn{1}{c|}{86.90}          & \multicolumn{1}{c|}{95.24}          & 91.07          \\
\multicolumn{1}{c|}{CADA~\cite{liu2022cada}}          & \multicolumn{1}{c|}{64.04}          & \multicolumn{1}{c|}{76.64}          & 70.34          \\
\multicolumn{1}{c|}{TAU~\cite{zhang2021tau}}           & \multicolumn{1}{c|}{54.20}          & \multicolumn{1}{c|}{78.60}          & 66.40          \\
\multicolumn{1}{c|}{ECSD-Net~\cite{liu2022ecsd}}      & \multicolumn{1}{c|}{80.20}          & \multicolumn{1}{c|}{86.50}          & 83.35          \\
\multicolumn{1}{c|}{BEAL~\cite{wang2019boundary}}      & \multicolumn{1}{c|}{81.00}          & \multicolumn{1}{c|}{89.80}          & 85.40        \\
\multicolumn{1}{c|}{DPL~\cite{Chen2021SourceFreeDA}}      & \multicolumn{1}{c|}{79.78}          & \multicolumn{1}{c|}{90.13}          &   84.96       \\
\multicolumn{1}{c|}{pOSAL~\cite{Wang2019}}        & \multicolumn{1}{c|}{78.70}          & \multicolumn{1}{c|}{86.50}          &    82.60     \\
\multicolumn{1}{c|}{Feng et al.~\cite{Feng2022}}        & \multicolumn{1}{c|}{84.10}          & \multicolumn{1}{c|}{90.50}          &    87.30     \\
\multicolumn{1}{c|}{Ours*} & \multicolumn{1}{c|}{83.47} & \multicolumn{1}{c|}{87.85} & 85.66 \\ \hline
\multicolumn{4}{c}{\textbf{Drishti-GS}}                                                                                         \\ \hline
% \multicolumn{1}{c|}{S-CUDA~\cite{liu2021s}}        & \multicolumn{1}{c|}{85.90}          & \multicolumn{1}{c|}{96.10}          & 91.00          \\
% \multicolumn{1}{c|}{NENet~\cite{pachade2021nenet}}         & \multicolumn{1}{c|}{84.01}          & \multicolumn{1}{c|}{96.32}          & 90.17          \\
\multicolumn{1}{c|}{CADA~\cite{liu2022cada}}          & \multicolumn{1}{c|}{84.00}          & \multicolumn{1}{c|}{89.00}          & 86.50           \\
\multicolumn{1}{c|}{TAU~\cite{zhang2021tau}}           & \multicolumn{1}{c|}{61.00}          & \multicolumn{1}{c|}{88.50}          & 74.75          \\
\multicolumn{1}{c|}{ECSD-Net~\cite{liu2022ecsd}}      & \multicolumn{1}{c|}{87.60}          & \multicolumn{1}{c|}{96.50}          & 92.05          \\
\multicolumn{1}{c|}{BEAL~\cite{wang2019boundary}}      & \multicolumn{1}{c|}{86.20}          & \multicolumn{1}{c|}{96.10}          &   91.15       \\
\multicolumn{1}{c|}{DPL~\cite{Chen2021SourceFreeDA}}      & \multicolumn{1}{c|}{83.53}          & \multicolumn{1}{c|}{96.39}          & 89.96          \\
\multicolumn{1}{c|}{pOSAL~\cite{Wang2019}}        & \multicolumn{1}{c|}{85.80}          & \multicolumn{1}{c|}{96.50}          &    91.15     \\
\multicolumn{1}{c|}{Feng et al.~\cite{Feng2022}}        & \multicolumn{1}{c|}{89.20}          & \multicolumn{1}{c|}{96.60}          &    92.90     \\
\multicolumn{1}{c|}{Ours*} & \multicolumn{1}{c|}{86.68} & \multicolumn{1}{c|}{96.17} & 91.43 \\ \hline
\end{tabular}
\caption{Comparing our method to other fundus segmentation methods. the results of other methods are obtained by using 400 fully labeled source domain images, whereas our method only
uses 40 source images. The training and testing images in the target domain used by all comparison methods are the same. The results for other methods are directly referenced from the articles. *Our method can achieve comparable and even better UDA segmentation performance only using $10 \%$ labeled source data, which indicates the
effectiveness of our method, as well as maintaining the
data efficiency.}
\end{table}

\begin{table}[!htbp]
\centering
\resizebox{11.7cm}{!}{
\label{table1}
\begin{tabular}{c|c|c|c|c|c|c|c|c} 
\hline
\multirow{2}{*}{\begin{tabular}[c]{@{}c@{}}\\Methods\end{tabular}} & \multicolumn{3}{c|}{Dice Metric [\%]}            & \multicolumn{3}{c|}{ASD Metric [pixel]}        & \multirow{2}{*}{\begin{tabular}[c]{@{}c@{}}Training \\Time [s/iter]\end{tabular}} & \multirow{2}{*}{\begin{tabular}[c]{@{}c@{}}Model \\Size [M]\end{tabular}}  \\ 
\cline{2-7}
                                                                   & Cup            & Disc           & Average        & Cup           & Disc           & Average       &                                                                                   &                                                                            \\ 
\hline
\multicolumn{9}{c}{\textbf{RIM-ONE-r3}}                                                                                                                                                                                                                                                                                                 \\ 
\hline
CyCADA                                                   & 66.61          & 76.99          & 71.80          & 47.35         & 41.62          & 44.48         & 12.14                                                                             & 31.04                                                                      \\
ADVENT                                                             & 67.99          & 80.67          & 74.33          & 42.04         & 33.43          & 37.74         & 12.19                                                                             & 42.61                                                                      \\
PixMatch                                                           & 70.50          & 75.20          & 72.85          & 16.33         & 35.90          & 26.12         & 13.26                                                                             & 42.61                                                                      \\
LTIR                                                               & 69.28          & 79.82          & 74.55          & 15.52         & 27.10          & 21.31         & 11.04                                                                             & 28.91                                                                      \\
MT                                                                 & 70.04          & 82.66          & 76.35          & 13.23         & 20.54          & 16.88         & 8.23                                                                              & 31.04                                                                      \\
PCS                                                                & 65.71          & 78.00          & 71.86          & 18.04         & 26.09          & 22.06         & 14.13                                                                             & 59.34                                                                      \\
\textbf{Ours}                                                      & \textbf{83.47} & \textbf{87.85} & \textbf{85.66} & \textbf{7.33} & \textbf{11.33} & \textbf{8.64} & 79.91                                                                             & 7.62                                                                       \\ 
\hline
\multicolumn{9}{c}{\textbf{Drishti-GS}}                                                                                                                                                                                                                                                                                                 \\ 
\hline
CyCADA                                                      & 81.83          & 91.54          & 86.68          & 12.55         & 12.32          & 12.43         & 7.71                                                                              & 31.04                                                                      \\
ADVENT                                                             & 81.82          & 92.32          & 87.07          & 12.43         & 10.59          & 15.25         & 7.33                                                                              & 42.61                                                                      \\
PixMatch                                                           & 75.31          & 93.13          & 84.22          & 16.91         & 8.34           & 12.63         & 8.71                                                                              & 42.61                                                                      \\
LTIR                                                               & 76.72          & 94.17          & 85.44          & 15.82         & 7.20           & 11.51         & 5.94                                                                              & 42.61                                                                      \\
MT                                                                 & 75.33          & 91.62          & 83.48          & 16.53         & 9.79           & 13.16         & 8.29                                                                              & 31.04                                                                      \\
PCS                                                                & 78.67          & 89.63          & 84.15          & 17.09         & 13.64          & 15.36         & 8.26                                                                              & 59.34                                                                       \\
\textbf{Ours}                                                      & \textbf{86.68} & \textbf{96.17} & \textbf{91.43} & \textbf{8.85} & \textbf{4.35}  & \textbf{6.60} & 86.43                                                                             & 7.62                                                                       \\
\hline
\end{tabular}
}
\caption{Experimental results of different domain adaptation approaches in terms of Dice and ASD metrics on RIM-ONE-r3 and Drishti-GS target datasets with 40 ($10\%$) labeled source data. Models' training time and models' sizes are also shown.}
\end{table}

\newpage
\bibliography{egbib}
\end{document}